\documentclass[acmtog,authorversion,nonacm]{acmart}

\usepackage{booktabs} 
\usepackage{stmaryrd}
\usepackage[normalem]{ulem}
\usepackage{multirow}

\usepackage[ruled]{algorithm2e} 

\SetAlFnt{\small}
\SetAlCapFnt{\small}
\SetAlCapNameFnt{\small}
\SetAlCapHSkip{0pt}

\acmJournal{TOG}

\setcopyright{rightsretained}

\newcommand{\revadd}[1]{{#1}}
\newcommand{\revdel}[1]{}

\newcommand{\vect}[1]{{\bf #1}}

\newcommand{\Lmae}{\mathcal{L}_\mathrm{1}}

\newcommand{\Lvgg}{\mathcal{L}_\mathrm{VGG}}
\newcommand{\Lstyle}{\mathcal{L}_\mathrm{style}}
\newcommand{\vggfeat}{\phi}
\newcommand{\maeWeight}{\lambda_{\mathrm{1}}}
\newcommand{\vggWeight}{\lambda_{\mathrm{vgg}}}
\newcommand{\styleWeight}{\lambda_{\mathrm{style}}}

\begin{document}

\acmJournal{TOG}
\acmYear{2023} 
\acmDOI{10.1145/3618393}
\title{ReShader: View-Dependent Highlights for Single Image View-Synthesis}

\author{Avinash Paliwal}
\orcid{1234-5678-9012-3456}
\affiliation{%
 \institution{Texas A\&M University}
 \city{College Station}
 \state{TX}
 \country{USA}}
\email{avinashpaliwal@tamu.edu}
\author{Brandon G. Nguyen}
\affiliation{%
 \institution{Texas A\&M University}
 \city{College Station}
 \state{TX}
 \country{USA}}
\email{bgn@tamu.edu}
\author{Andrii Tsarov}
\affiliation{%
\institution{Leia Inc.}
\streetaddress{Rono-Hills}
\city{Menlo Park}
\state{CA}
\country{USA}}
\email{andrii.tsarov@leiainc.com}
\author{Nima Khademi Kalantari}
\affiliation{%
 \institution{Texas A\&M University}
 \city{College Station}
 \state{TX}
 \country{USA}}
\email{nimak@tamu.edu}

\begin{abstract}
In recent years, novel view synthesis from a single image has seen significant progress thanks to the rapid advancements in 3D scene representation and image inpainting techniques. While the current approaches are able to synthesize geometrically consistent novel views, they often do not handle the view-dependent effects properly. Specifically, the highlights in their synthesized images usually appear to be glued to the surfaces, making the novel views unrealistic. To address this major problem, we make a key observation that the process of synthesizing novel views requires changing the shading of the pixels based on the novel camera, and moving them to appropriate locations. Therefore, we propose to split the view synthesis process into two independent tasks of pixel reshading and relocation. During the reshading process, we take the single image as the input and adjust its shading based on the novel camera. This reshaded image is then used as the input to an existing view synthesis method to relocate the pixels and produce the final novel view image. We propose to use a neural network to perform reshading and generate a large set of synthetic input-reshaded pairs to train our network. We demonstrate that our approach produces plausible novel view images with realistic moving highlights on a variety of real world scenes.

\end{abstract}

\begin{teaserfigure}
  \includegraphics[width=\textwidth]{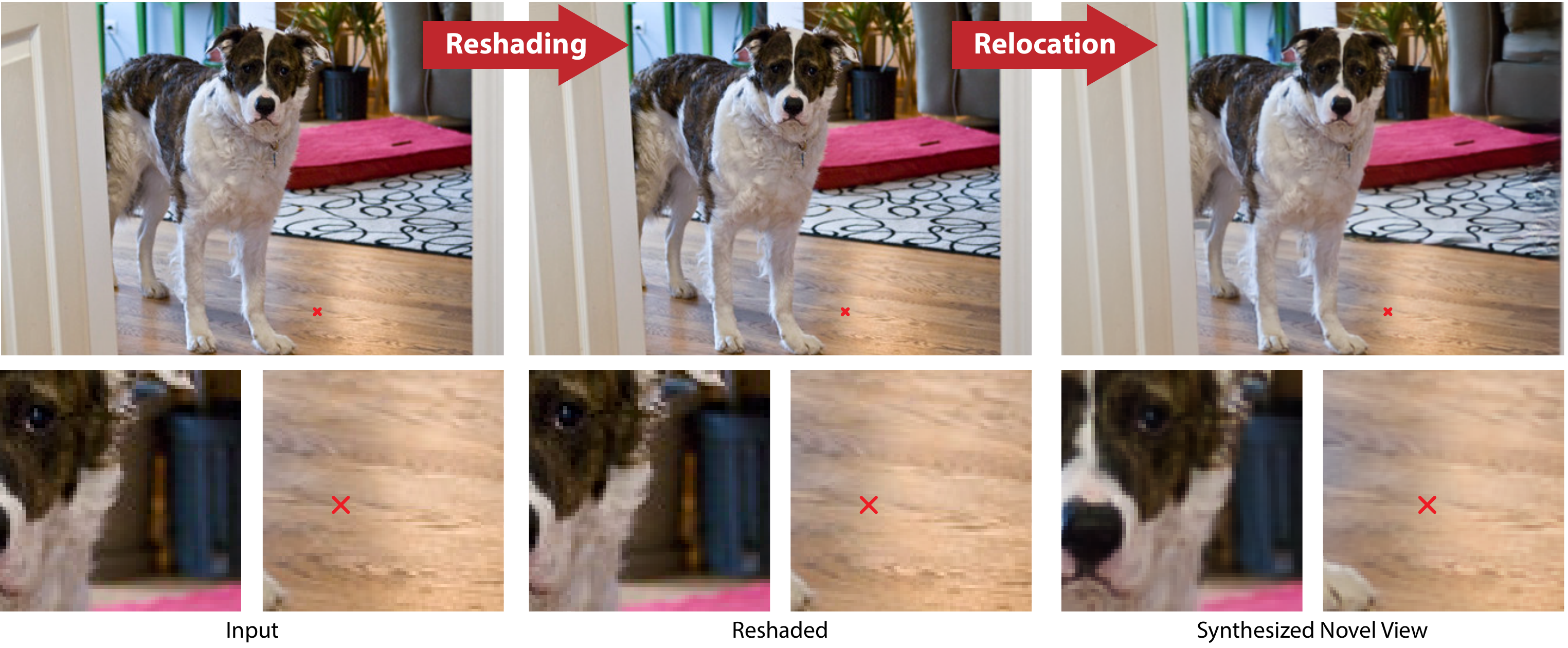}
  \vspace{-0.25in}
  \caption{To properly handle the view-dependent effects, we propose to break down the view synthesis process into two tasks of pixel reshading and relocation. During reshading, we use a neural network to generate a new version of the input image (shown on the left) with the shading computed based on the novel view. As shown on the middle, our reshading network correctly leaves the diffuse areas intact (the dog's head), but moves the highlights on the specular areas (wooden floor). The relocation process takes this reshaded image and generates the novel view image. The red crosses mark the same location on the wooden floor to make it easier to observe the effect of reshading and relocation.}
  \Description{This is the teaser figure for the article.}
  \label{fig:teaser}
\end{teaserfigure}

%



\maketitle
\section{Introduction}

Creating novel views of a scene from a single image is a compelling way to breathe life into still photographs. When displayed on virtual reality (e.g., HTC vive and Meta Quest) or light field (e.g., Lume Pad~\cite{Leia2023LumePad}) devices, these ``3D photographs'' provide a highly immersive experience for users, allowing them to vividly relive moments captured in still photographs as if they have been transported back in time and place.

\begin{figure}
  \includegraphics[width=\linewidth]{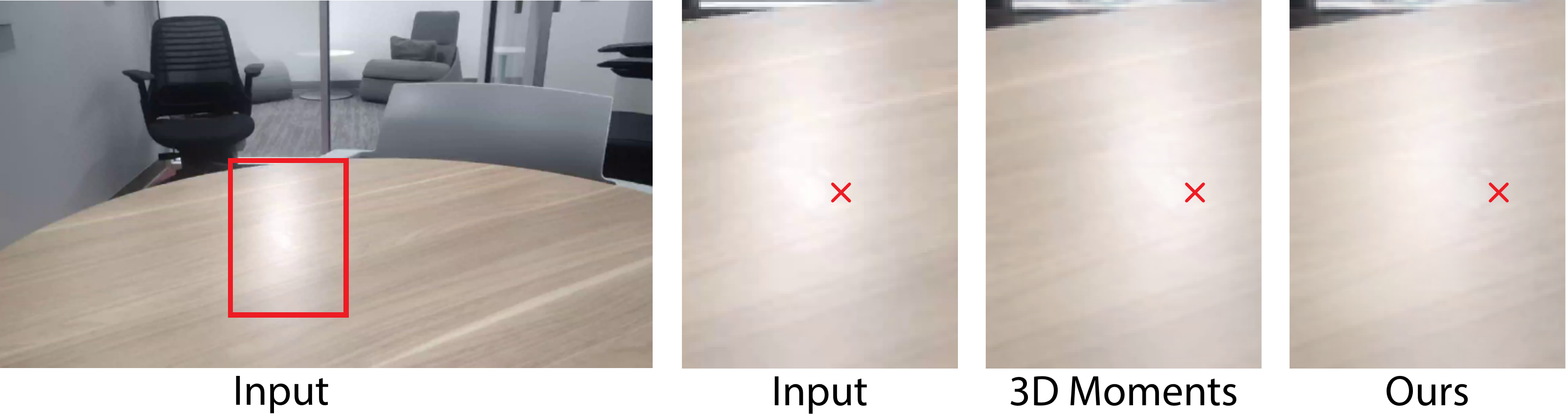}
  \vspace{-0.3in}
  \caption{We compare our results against 3D Moments by Wang et al.~\shortcite{wang2022_3dmoments}. 3D Moments reconstructs the novel image by moving the input pixels according to their depth values. As such, the highlights are treated as textures and appear to be glued to the wooden table. Our approach, however, is able to properly move the highlights over the table. The red crosses mark the same location on the table. Note that the cross is inside the highlight in the input and 3D Moment's results, but it appears to be outside the highlight in our results.}
  \label{fig:motivation}
  \vspace{-0.2in}
\end{figure}

The rapid advancements in 3D scene representation and image inpainting techniques have led to remarkable progress in single image view synthesis in recent years. Despite this, the existing techniques focus on producing geometrically consistent novel views and mostly ignore the view-dependent effects. For example, a number of techniques~\cite{Shih3DP20, jampani:ICCV:2021}, handle this application in a \emph{modular} manner. These approaches estimate the depth from the input and use it to decompose the scene into multiple layers. These depth layers are then warped to the novel view and composed together to form the final image. Unfortunately, these methods treat the highlights, which are quite common in real scenes, as textures and warp them to the novel views along with other areas. Therefore, as shown in Fig.~\ref{fig:motivation}, the highlights in their synthesized views appear to be glued to the surfaces, making their results unrealistic.

On the other hand, several approaches~\cite{srinivasan2017learning,Li2020LF,yu2021pixelnerf} handle this problem by learning the process in an \emph{end-to-end} manner. These techniques learn the entire view synthesis pipeline either directly~\cite{srinivasan2017learning}, or through various scene representations, such as neural radiance field (NeRF)~\cite{yu2021pixelnerf} and multiplane images (MPI)~\cite{Li2020LF,tucker2020mpi}. Although they could potentially handle the view-dependent effects, these techniques often struggle to properly reconstruct the moving highlights.

Our main observation is that both the shading and projected pixel location of a 3D surface point change between the input and novel view images. Modular approaches overlook the view-dependent shading, focusing solely on pixel relocation. The end-to-end approaches, on the other hand, aim to learn to move the pixels and change their shading within a unified system. However, the majority of effort is dedicated to learning pixel relocation, as the contribution of the shading mismatch to their training loss is often minimal.

Guided by this observation, we make a key contribution to break down the novel view synthesis process into two tasks: pixel reshading and relocation (see Fig.~\ref{fig:teaser}). During the reshading process, we only adjust the shading of the input image according to the novel camera. We then perform pixel relocation on the reshaded image, using the modular method by Wang et al.~\shortcite{wang2022_3dmoments}, to obtain the final novel view image. We propose to learn the reshading process using a neural network that takes a single image as well as the relative novel camera position as the input and produces the reshaded image. Since there are no publicly available datasets of input-reshaded image pairs, we render a large number of synthetic image pairs for training. We train our reshading network on this newly introduced dataset using a perceptual loss to ensure producing plausible, but detailed reshaded images. We demonstrate that our method produces high-quality novel view images with plausible moving highlights on a wide range of real scenes.

\section{Related Work}

The problem of view synthesis has been extensively studied and many powerful multi- and single-image methods have been developed~\cite{mildenhall2020nerf,Wizadwongsa2021NeX,Shih3DP20,tucker2020mpi}. A complete literature review is beyond the scope of this paper. Here, we mainly focus on approaches that use a single image as the input. We also discuss image relighting methods as they are relevant to the focus of our paper.

\subsection{Single Image View Synthesis}

We discuss these approaches by categorizing them into two classes of modular and end-to-end. The modular methods~\cite{Niklaus_2019_KenBurn,Kopf_2019_CVPRW,Kopf_2020_TOG,Shih3DP20,jampani:ICCV:2021,wang2022_3dmoments} break down the process into multiple components and address each component separately. Specifically, these techniques divide the view synthesis pipeline into depth estimation, image warping, and image inpainting. The individual methods differ in how they handle each stage of the pipeline. For example, Niklaus et al.~\shortcite{Niklaus_2019_KenBurn} train a depth estimation network and use it to directly reproject the input image to the novel view. On the other hand, Shih et al.~\shortcite{Shih3DP20} obtain the depth using an existing method~\cite{Ranftl2022} and reconstructs layered depth image (LDI) representation~\cite{shade1998layereddepth} to warp the input image to the novel view. These techniques, however, primarily focus on pixel relocation and overlook the pixel reshading process. As a result, they produce results with incorrect view-dependent effects, where the highlights appear to be glued to the surfaces (see Fig.~\ref{fig:motivation}).

\revadd{A category of modular methods focus on handling the view-dependent effects by first decomposing the image(s) into multiple layers (e.g., diffuse and reflective), warping each layer separately, and blending them to generate the final image. However, most of these techniques are either specifically designed for rendering~\cite{Lochmann2014real,Zimmer2015path} where ground truth scene information (e.g., geometry and material) is available, or require multiple images~\cite{Blake1985_IJCAI,Roth2006specular,Sinha2012image}.} 

In contrast \revadd{to the modular approaches}, a number of techniques~\cite{srinivasan2017learning,wiles2020synsin,Li2020LF,tucker2020mpi,Han2022AdaptiveMPI,yu2021pixelnerf} attempt to learn the entire view synthesis process in an end-to-end manner. Zhou et al.~\shortcite{Zhou_2016_ECCV} propose to estimate optical flows at novel views and use the estimated flow to backward warp the input image. The flow estimation network is trained by minimizing the loss between the synthesized and ground truth novel view images. Srinivasan et al.~\shortcite{srinivasan2017learning} propose to estimate a light field from a single image using a convolutional neural network (CNN). Several approaches use a network to estimate intermediate representations, such as point cloud~\cite{wiles2020synsin}, multiplane images (MPI)~\cite{Li2020LF,tucker2020mpi,Han2022AdaptiveMPI}, and neural radiance field (NeRF)~\cite{yu2021pixelnerf}. Since these approaches perform end-to-end training, they could potentially learn to handle the view-dependent effects. However, highlights are usually concentrated in small regions, and thus the shading mismatch does not significantly contribute to the loss function. As such, these methods often are not able to produce results with proper moving highlights.

Recently, several approaches~\cite{poole2022dreamfusion,gu2023nerfdiff,shue20223d,watson2022novel,fridman2023scenescape,chan2023genvs} have proposed to address this problem using diffusion models~\cite{ho2020denoising}. Some of these techniques~\cite{shue20223d, watson2022novel} produce novel view images of only single objects or simple scenes. Others~\cite{fridman2023scenescape, chan2023genvs} handle complex scenes and produce impressive walkthroughs from a single image. However, when synthesizing views that are relatively close to the input, the quality of their synthesized images are not on par with the existing modular or MPI-based techniques.

\subsection{Image Relighting}

Image relighting is the process of reconstructing images of a scene under different illumination. This problem is highly related to inverse rendering where the aim is to estimate the image formation factors (e.g., shape, reflectance, lighting) of a scene. Several methods propose to handle this application either by directly estimating the relit images~\cite{xu2018deep}, estimating the individual factors~\cite{xu2019deep}, or by utilizing NeRF~\cite{bi2020deep, bi2020neural, boss2021nerd, srinivasan2021nerv,zhang2021nerfactor}. However, these approaches focus on simple scenes or single objects, and require multiple images as the input. For more complex scenes, Philip et al.~\shortcite{Philip_2019_TOG} propose a relighting approach for outdoor scenes, while Philip et al.~\shortcite{Philip2021RelightMVS} and Wu et al.~\shortcite{wu2022scalable} focus on indoor scenes. However, both of these techniques use several images of the scene as the input.

Several techniques~\cite{Sengupta_2019_ICCV,Li_2020_CVPR,wang2021learning,li2022physically} propose to estimate all the image formation factors including shape, reflectance, and lighting, from a single image. Sengupta et al.~\shortcite{Sengupta_2019_ICCV} propose an inverse rendering network to estimate albedo, normal, and a single environment lighting. Li et al.~\shortcite{Li_2020_CVPR} extend this work to estimate per-pixel lighting, as well as roughness and depth. Wang et al.~\shortcite{wang2021learning} further propose to estimate 3D lighting of the scene through volumetric spherical Gaussian. Moreover, Li et al.~\shortcite{li2022physically} present a holistic scene reconstruction system that estimates the reflectance, shape, and parameteric 3D lighting. These techniques demonstrate impressive results for object insertion, material editing, and dramatic lighting change~\cite{li2022physically} (e.g., covering a window). While they could potentially be used to perform pixel reshading, these methods do not meet the quality requirement for our application.  

\section{Algorithm}

Given a single RGB image $I$, captured with a camera at location $\vect{c}$, our primary goal is to synthesize an image $I^{\prime}$ from a novel view $\vect{c}^\prime$. Similar to most existing methods~\cite{jampani:ICCV:2021,Han2022AdaptiveMPI}, we assume the depth can be obtained with a reasonable accuracy using single image depth estimation techniques~\cite{Ranftl2022}.

We begin by discussing the rendering equation~\cite{kajiya1986rendering}, a reasonably expressive rendering model, to describe the relationship between the input and novel view images. Formally, the rendering equation describes the total outgoing radiance $L_o(\vect{x}, \vect{\omega}_o)$ at a 3D point $\vect{x}$ along the viewing direction $\omega_o$ as follows:

\vspace{-0.1in}
\begin{equation}
\label{eqn:rendering}
L_o(\vect{x}, \vect{\omega}_o) = L_e(\vect{x}, \vect{\omega}_o) + \int_{\Omega} f_r(\vect{x}, \vect{\omega}_o, \vect{\omega}_i) \ L_i(\vect{x}, \vect{\omega}_i) \ \cos(\theta_i) \ d\vect{\omega}_i,
\end{equation}

\noindent where $L_e$ and $L_i$ are the emitted and incoming radiances, respectively, $\vect{\omega}_i$ is the incoming direction, and $f_r$ is the bidirectional reflectance distribution function (BRDF). Moreover, $\theta_i$ is the angle between $\vect{\omega}_i$ and the surface normal, and the integral is taken over the entire hemisphere $\Omega$ over the surface point.

\begin{figure}
  \includegraphics[width=\linewidth]{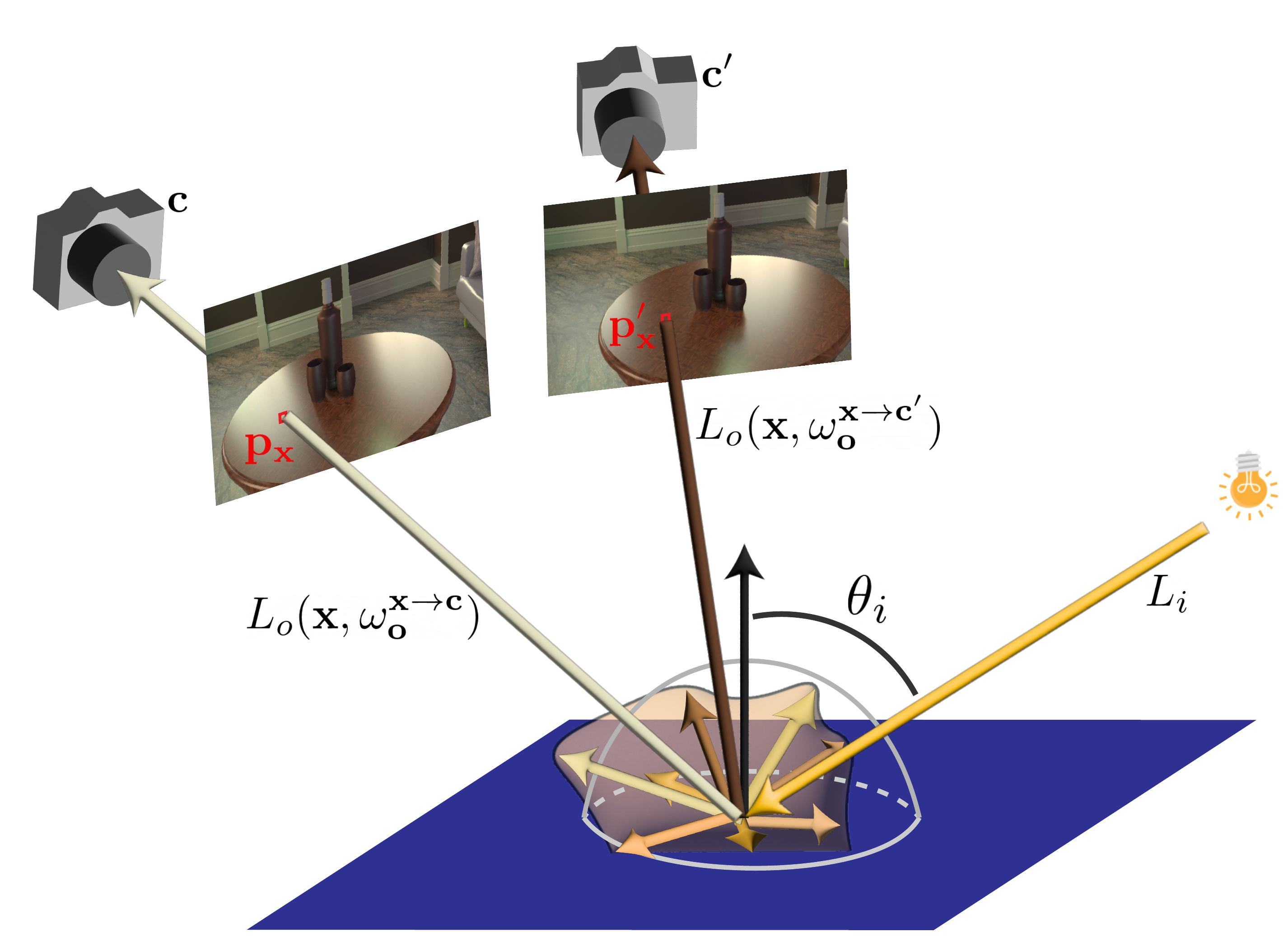}
  \vspace{-0.2in}
  \caption{We visualize the image formation process for the input ($\vect{c}$) and novel ($\vect{c}^\prime$) cameras. A surface point $\vect{x}$ appears at two different locations ($\vect{p}_{\vect{x}}$ and $\vect{p}^\prime_{\vect{x}}$) in the input and novel images. Moreover, the shading of point $\vect{x}$ in the two images is determined by $L_o(\vect{x}, \vect{\omega}_o^{\vect{x}\shortrightarrow \vect{c}})$ and $L_o(\vect{x}, \vect{\omega}_o^{\vect{x}\shortrightarrow \vect{c}^\prime})$, and thus is different. Note that the incoming radiance $L_i$, surface normal (and consequently $\theta_i$), and the BRDF (shown with curly black line), are the same for both the input and novel view images.}
  \label{fig:reflection}
  \vspace{-0.15in}
\end{figure}

\begin{figure}
  \includegraphics[width=\linewidth]{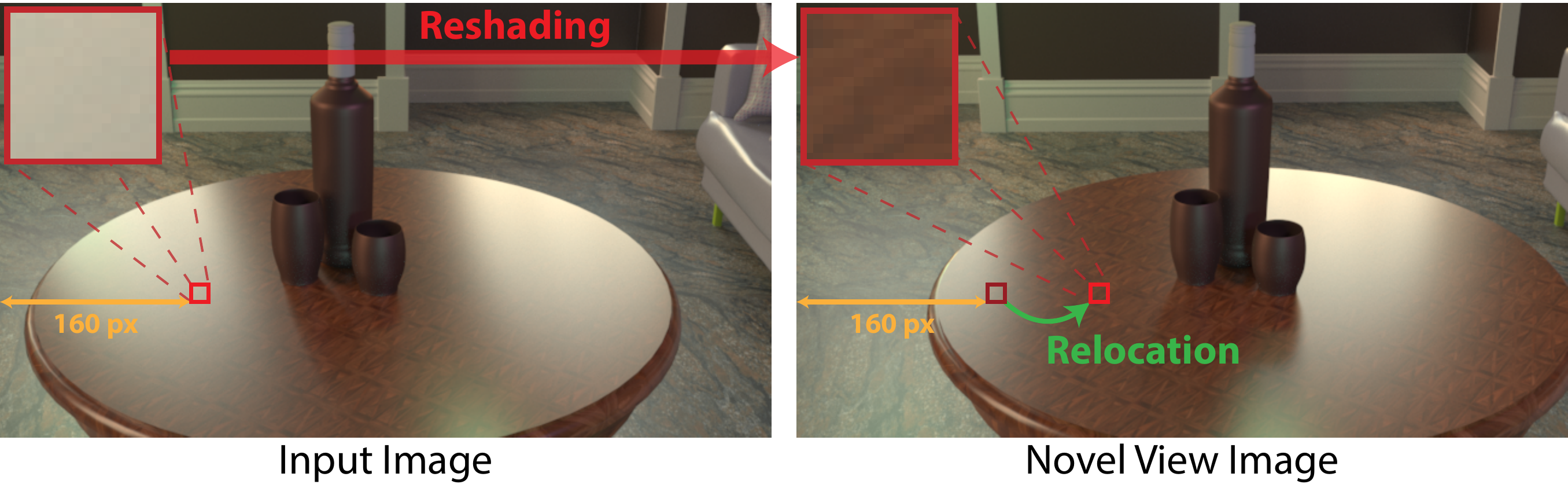}
  \vspace{-0.25in}
  \caption{We show an input and a novel view image. The same point on the table appears at different locations and with different shadings in the input and novel images. Therefore, the view synthesis process can be divided into two tasks of pixel reshading and relocation.}
  \label{fig:warping}
  \vspace{-0.2in}
\end{figure}

As shown in Fig.~\ref{fig:reflection}, the appearance of a surface point $\vect{x}$ in the input and novel images is determined by the outgoing radiance $L_o(\vect{x}, \vect{\omega}_o^{\vect{x}\shortrightarrow \vect{c}})$ and $L_o(\vect{x}, \vect{\omega}_o^{\vect{x}\shortrightarrow \vect{c}^\prime})$, respectively. Here, $\vect{\omega}_o^{\vect{x}\shortrightarrow \vect{c}}$ is the direction from the surface point to input camera location $\vect{c}$, while $\vect{\omega}_o^{\vect{x}\shortrightarrow \vect{c}^\prime}$ represents the direction to the novel camera at position $\vect{c}^\prime$.

Based on this analysis, we observe that the appearance of point $\vect{x}$ in the input and novel images differs in two major ways: {\bf 1)} The point $\vect{x}$ appears with different shadings in the input and novel images as its appearance is determined by $L_o(\vect{x}, \vect{\omega}_o^{\vect{x}\shortrightarrow \vect{c}})$ and $L_o(\vect{x}, \vect{\omega}_o^{\vect{x}\shortrightarrow \vect{c}^\prime})$, respectively. {\bf 2)} The location of this point in the two images is different; $\vect{p}_{\vect{x}}$ and $\vect{p}^\prime_{\vect{x}}$ in the input and novel images, respectively. This is determined by the intersection of the rays along directions $\vect{\omega}_o^{\vect{x}\shortrightarrow \vect{c}}$ and $\vect{\omega}_o^{\vect{x}\shortrightarrow \vect{c}^\prime}$ with the image planes of the input and novel cameras, respectively.

Therefore, we can describe the view synthesis process through two tasks of pixel reshading and relocation, as shown in Fig.~\ref{fig:warping}. Existing modular approaches~\cite{Shih3DP20, jampani:ICCV:2021}, synthesize novel view images by warping the input image to the novel view using the input depth. As such, they mainly focus on the pixel relocation task and ignore the pixel reshading process, which is responsible for the view-dependent effects. The end-to-end systems~\cite{Li2020LF, Han2022AdaptiveMPI}, on the other hand, attempt to learn both pixel reshading and relocation processes by minimizing the loss between the estimated and ground truth novel view images. However, these systems often ignore the pixel reshading task as the contribution of the shading differences to the appearance loss is small; view-dependent highlight are often concentrated in small regions in each scene. As such, these techniques are not able to properly handle the view-dependent effects.

To address this problem, we propose to treat pixel reshading and relocation as two independent tasks. Specifically, we first adjust the shading of the input image according to the novel view camera. We then use the reshaded image as the input to the approach by Wang et al.~\shortcite{wang2022_3dmoments} to relocate the pixels and produce the final image. Below we discuss our approach in detail.

\subsection{Pixel Reshading}

Our goal is to take the input image $I$ and produce a reshaded image $I_s$ that has the same shading as the novel view image. This necessitates changing the shading of input pixel $\vect{p}_{\vect{x}}$ from $L_o(\vect{x}, \vect{\omega}_o^{\vect{x}\shortrightarrow \vect{c}})$, to the shading of the corresponding pixel in the novel image $\vect{p}^{\prime}_{\vect{x}}$, i.e., $L_o(\vect{x}, \vect{\omega}_o^{\vect{x}\shortrightarrow \vect{c}^\prime})$. Note that at this stage, we are not interested in pixel relocation and reshading occurs in the input camera frame.

According to the rendering equation (Eq.~\ref{eqn:rendering}), performing the reshading process requires estimating various components: the lighting $L_e$ (emitters), material properties $f_r$, incoming radiance from all directions going through the hemisphere $L_i$, and the normals (to compute $\theta_i$). Once these quantities are estimated, it is possible to recompute the shading of pixel $\vect{p}_{\vect{x}}$ in the input image, by evaluating the integral in Eq.~\ref{eqn:rendering} using the outgoing direction of the corresponding pixel in the novel view image $\vect{\omega}_o^{\vect{x}\shortrightarrow \vect{c}^\prime}$. Note that the outgoing direction can be easily inferred from the input depth and the camera positions (provided relatively to avoid the need for estimating the input camera pose).

Unfortunately, estimating all of the aforementioned quantities from a single image is an extremely challenging problem. While there are existing techniques~\cite{Sengupta_2019_ICCV,Li_2020_CVPR,wang2021learning,li2022physically} that estimate these various factors to a great extent, the quality of their re-rendered images falls short of the requirements for our view synthesis application.

Therefore, we instead propose to directly learn the reshaded image from the input image using a neural network. Although simple, as shown in Sec.~\ref{sec:results} and in the supplementary video, our method is able to handle this challenging problem reasonably well and produce results with plausible moving highlights. In the following sections, we describe our dataset, inputs, architecture, and training process.

\begin{figure}
  \includegraphics[width=\linewidth]{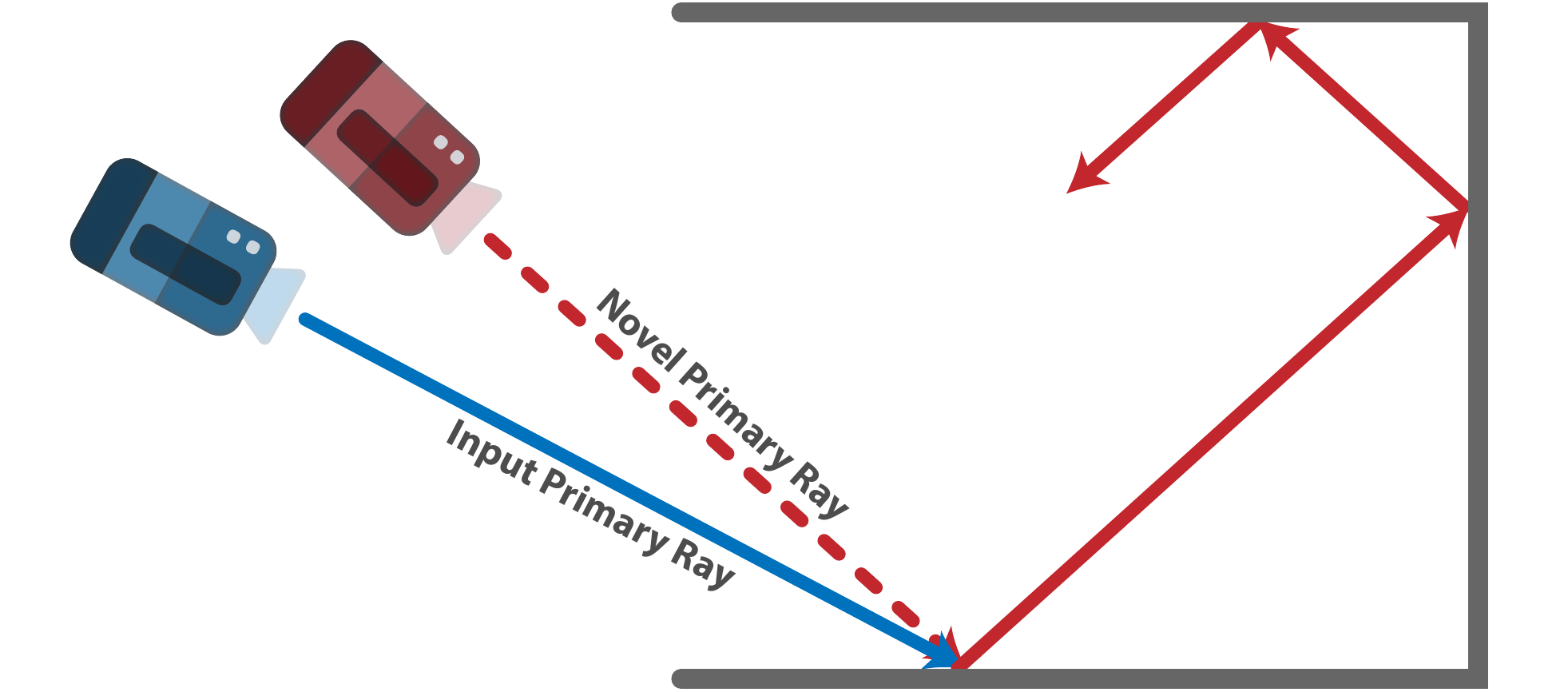}
  \caption{We visualize our modification to the path tracer to render the reshaded images. We trace a primary ray to find the first intersection from the input camera. We then find the ray from the novel camera to this point (novel primary ray). This ray is then used for shading computation at the intersection point and generation of the secondary ray.}
  \label{fig:rendering}
  \vspace{-0.15in}
\end{figure}

\begin{figure}
  \includegraphics[width=\linewidth]{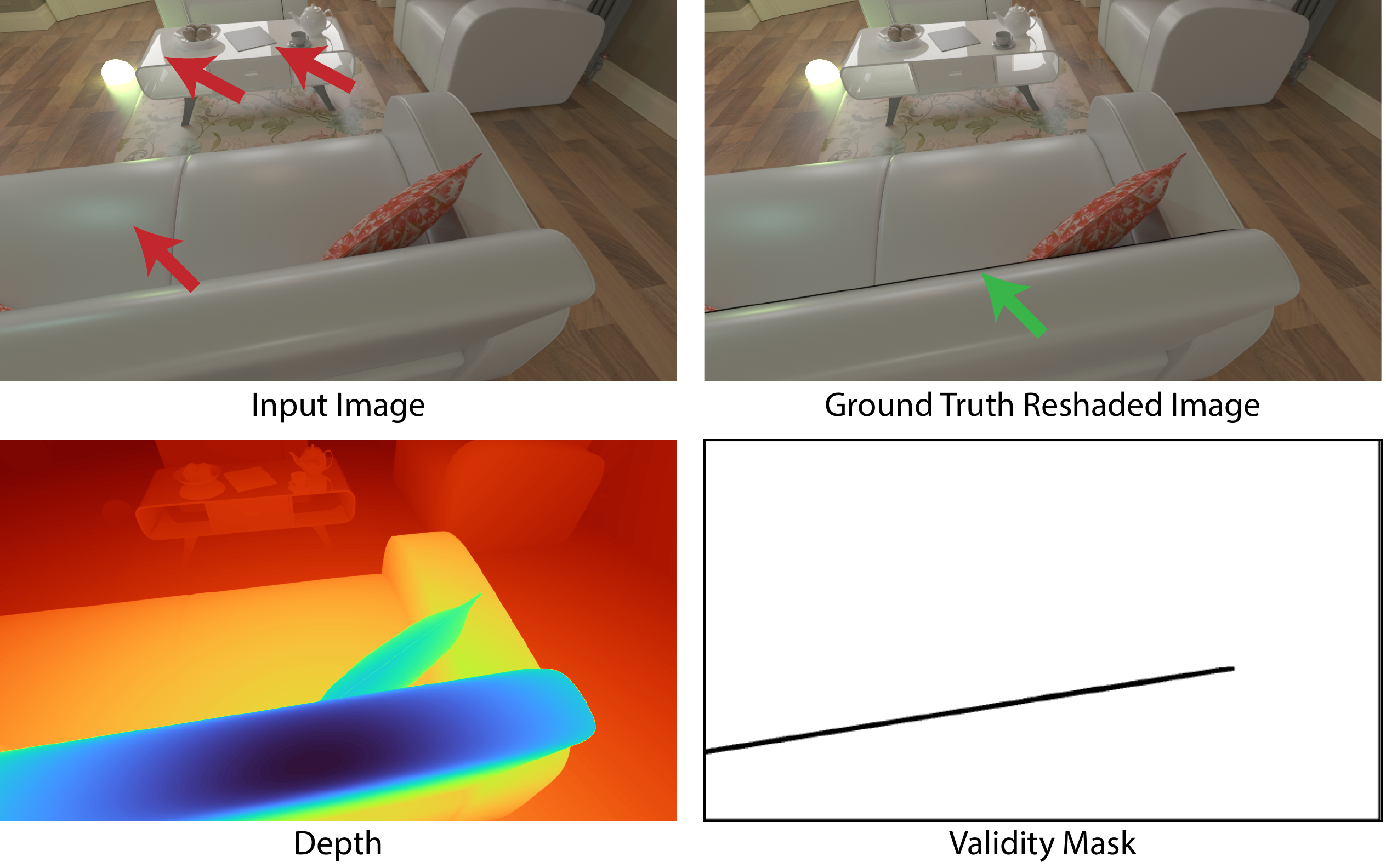}
  \vspace{-0.25in}
  \caption{For each training example in our dataset, we store the input and ground truth reshaded images, as well as the depth and validity mask. The red arrows point to the highlights in the input image that are moved in the reshaded image. Note that the objects in the reshaded image are in the same location as the input image, since reshading happens in the input camera frame. Small areas in the reshaded image (indicated by the green arrow) contain incorrect shading. We mask these out using the validity mask in our training loss.}
  \label{fig:inputs}
  \vspace{-0.2in}
\end{figure}

\subsection{Dataset}

To train our reshading network, we need a dataset of input-reshaded image pairs, which is currently not available. Unfortunately, obtaining such a dataset from real scenes is extremely challenging. Capturing the reshaded image necessitates taking a picture of the scene from the input camera view, but with the light rays going towards a different camera. One potential solution is to take a large number of images of a scene and use neural radiance field (NeRF)~\cite{mildenhall2020nerf} to reconstruct the radiance field of the scene. This radiance field can then be used to produce the reshaded images. However, generating a large scale dataset using this approach is difficult. Additionally, even the state-of-the-art approaches~\cite{kopanas2022neural,kerbl3Dgaussians,verbin2022refnerf} struggle to produce high-quality view-dependent effects on arbitrary surfaces.

Therefore, we propose to generate our input-reshaded image pairs synthetically. Specifically, we use the Tungsten renderer~\cite{Bitterli2014Tungsten} and render our input images using a large number of samples per pixel. We then slightly modify the path tracer to obtain the corresponding reshaded images, as shown in Fig.~\ref{fig:rendering}. To do this, we trace primary rays from the input camera (input primary ray) to find the first intersection points. We then calculate the rays connecting the novel camera to these intersection point (novel primary ray). These novel primary rays are then used for shading and generating all the additional secondary rays. An example input-reshaded image pair from our dataset is shown in Fig.~\ref{fig:inputs} (top row).

Note that some regions from the input image are occluded in the novel camera. We could easily detect and mask these areas by performing a visibility check with the novel primary ray. However, we choose not to do so to provide more content for our network to learn from. Most of the occluded areas will be shaded correctly as if they are not obscured from the camera. However, small regions (see the green arrow in Fig.~\ref{fig:inputs}), typically along the boundaries of objects, will be incorrectly shaded. These are the cases where the angle between the surface normal and novel primary ray is greater than 90 degrees. We detect these regions and create a validity mask, as shown in Fig.~\ref{fig:inputs}, which is used to mask out such areas when computing our training loss. Note that since we are using Monte Carlo rendering, each pixel is rendered by tracing a large number of rays. We mark a pixel as invalid if any of such rays does not satisfy our constraint. This is why the line in the validity mask appears to be thicker than the problematic region in the reshaded image.

\begin{figure}
  \includegraphics[width=\linewidth]{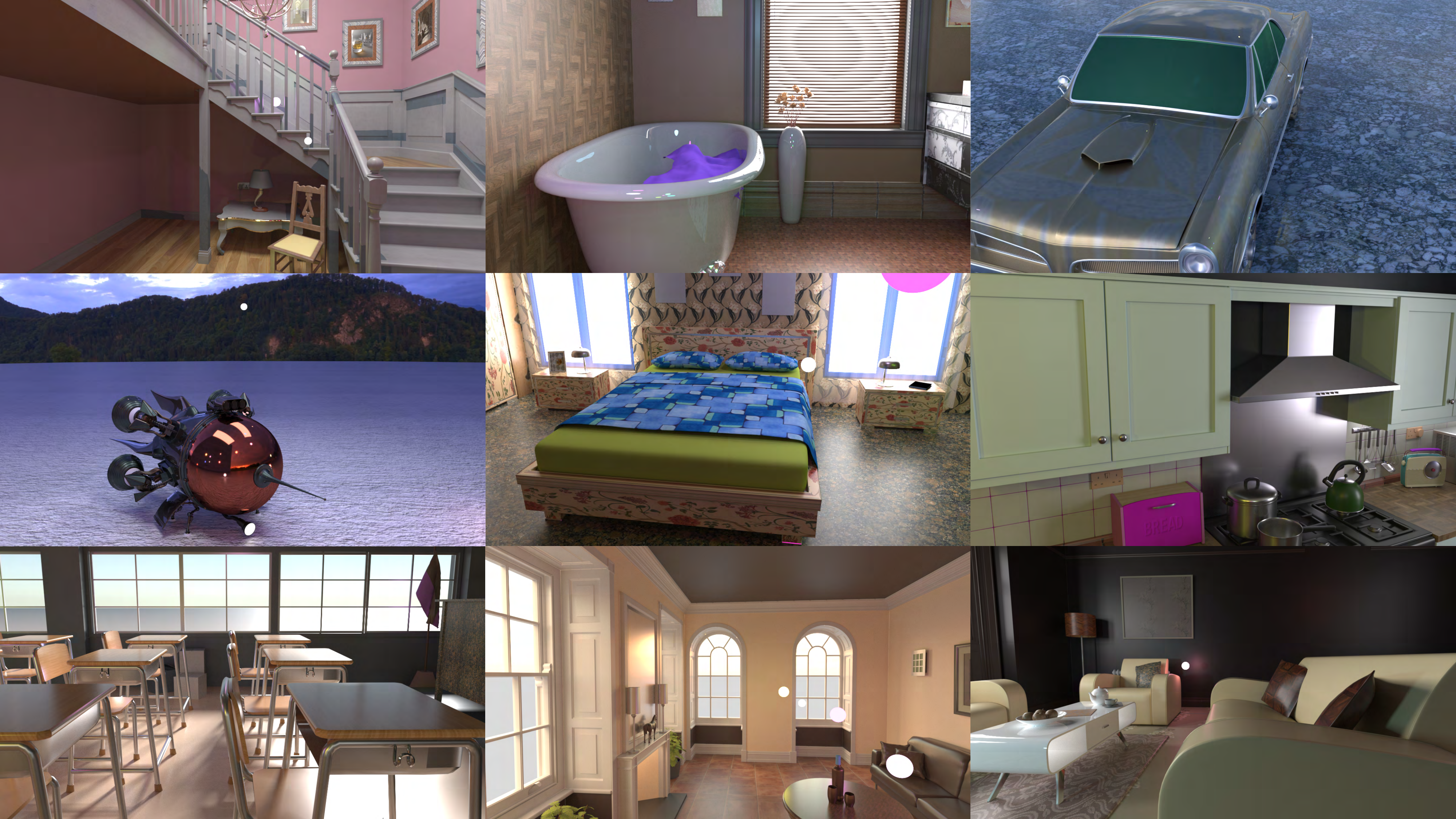}
  \vspace{-0.25in}
  \caption{Scenes used to create the synthetic dataset.}
  \label{fig:dataset}
  \vspace{-0.2in}
\end{figure}

We use the above approach to generate our synthetic dataset using 9 scenes, shown in Fig.~\ref{fig:dataset}, provided by Bitterli~\shortcite{resources16}. For each scene, we render 200 input-reshaded pairs by randomly placing the input and novel cameras inside the scene. We randomly choose the novel cameras inside a sphere, centered on the input camera, with radii ranging from 0.1 to 0.3. \revadd{Note that since all the scenes have similar global scale, the chosen radius range corresponds to a reasonable and uniform camera movement in all the training scenes.} For every image pair, we randomly change the texture and material properties of the objects in the scene. By default, most scenes only use the environment map as the light source. To increase the robustness of our approach, we add multiple random colored orbs into the scene at random locations. We render $1280\times720$ high dynamic range (HDR) images with 8K samples per pixel and for each example, we store the input and reshaded images, as well as the depth, validity mask, and the metadata of the cameras. Our training data for one example is shown in Fig.~\ref{fig:inputs}.

\begin{figure}
  \includegraphics[width=\linewidth]{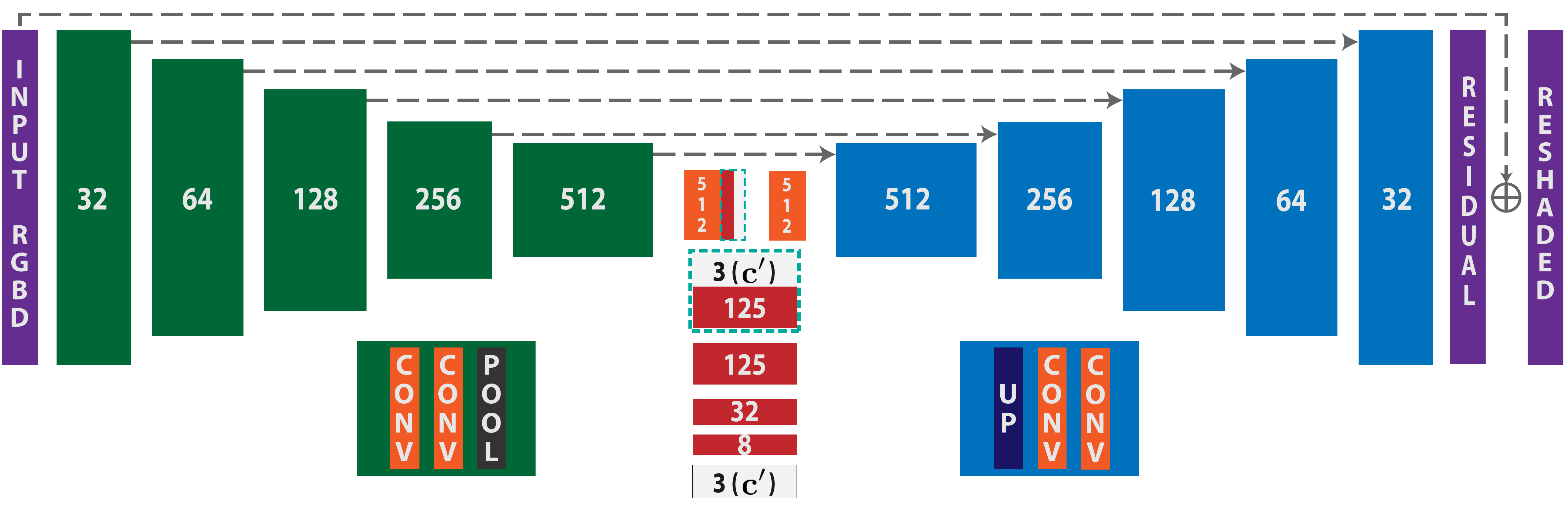}
  \vspace{-0.2in}
  \caption{We show the architecture of our reshading network. Each convolutional layers (shown in orange) is followed by a LeakyReLU activation, except the last layer that uses tanh activation. We use average $2\times 2$ pooling for downsampling, while we use bilinear upsampling to increase the resolution. We use an MLP to convert the 3 channel novel camera position vector to a 125-channel feature vector. We then concatenate the original camera position vector with this feature vector. The result is then replicated and attached to the bottleneck feature map. The dashed lines represent skip connections. Note that our network estimate the residual image which is added to the input to obtain the reshaded image.}
  \label{fig:architecture}
  \vspace{-0.2in}
\end{figure}

\subsection{Inputs}

For our network to be able to properly reshade an input image, we need to provide the depth information along with the novel camera position to our network. The novel camera position is a 3-channel vector containing position of the novel camera relative to the input camera. Similar to most current single image view synthesis methods, we estimate the depth map using an existing single image depth estimation method (Ranftl et al.'s approach~\shortcite{Ranftl2021,Ranftl2022} in our implementation). Instead of passing the depth to our network, however, we first convert it to disparity. We then scale it by a factor of 1/4 and clamp it to one. This ensures that the disparity is in the range [0, 1] and it covers the depth from 0.25 to infinity. Moreover, we apply frequency encoding~\cite{mildenhall2020nerf} with 5 frequencies (11 channels; original plus 5 sines and 5 cosines) to the input disparity to allow the network to effectively use the disparity, particularly for far away regions. Frequency encoding essentially increases the resolution of the disparity, while remaining in the range [0, 1]; similar disparity values will have significantly different representation in the frequency domain.

To summarize, we use the input RGB image, frequency encoded disparity map, and the relative novel camera position as the input to our network to produce the reshaded image. The effect of using the disparity map and frequency encoding are shown in Figs.~\ref{fig:ablation_depth}~and~\ref{fig:ablation_freq_enc}, respectively.  

\subsection{Architecture}

We utilize a UNet~\cite{ronneberger2015u} style encoder-decoder style architecture consisting of 5 downsampling/upsampling layers. The encoder takes the input image and frequency encoded disparity (3+11 channels) and produces a bottleneck feature map of size $H/32\times W/32\times 512$, where $H$ and $W$ are the height and width of the input image, respectively. The three channel novel camera position vector is converted to a 125-channel feature vector using a multilayer perceptron (MLP) with a series of fully connected layers. This feature vector is then concatenated with the original 3-channel camera position vector to produce our novel camera features. This is then replicated and concatenated with the bottleneck feature map from the encoder (map of size $H/32\times W/32\times 640$). The concatenated feature map is then used as the input to the decoder to produce a 3-channel residual image. The residual is then added to the input to produce the reshaded image. Our architecture is shown in Fig.~\ref{fig:architecture}.

\subsection{Training}

We perform a series of augmentations to improve the generalization ability of our network. We take $384\times384$ random crops of the HDR synthetic dataset and convert the input and ground truth reshaded pairs to low dynamic range images by applying random exposure (scale factor between 3 and 10) and gamma correction ($\gamma$ between 2.2 and 5). In addition, we randomly scale the disparity by a factor of $f$ and the camera position by a factor of $1/f$ simultaneously. This increases the range of scene scales in our training data.

Since this problem is highly ill-posed, we perform the training using a combination of $\mathcal{L}_1$ and perceptual losses. Specifically, our loss consists of the following three terms:

\vspace{-0.1in}
\begin{equation}
\label{eqn:total}
\mathcal{L} = \maeWeight\Lmae + \vggWeight\Lvgg + \styleWeight\Lstyle,
\end{equation}

\noindent where the first term is the $\mathcal{L}_1$ loss between the estimated and ground truth reshaded images and is defined as follows:

\begin{equation}
\label{eqn:l1}
\Lmae = \Vert \tilde{I}_s - I_s \Vert_1.
\end{equation}

Moreover, the second term is a perceptual VGG-based loss and is defined as: 

\vspace{-0.1in}
\begin{equation}
\label{eqn:vgg}
\Lvgg = \Vert \vggfeat(\tilde{I}_s) - \vggfeat(I_s) \Vert_2^2,
\end{equation}

\noindent where $\vggfeat$ represents the output features from the \texttt{conv4\_4} layer of VGG-19~\cite{simonyan2014very}. Furthermore, the third term is a perceptual VGG-based style loss and is defined as:

\vspace{-0.1in}
\begin{equation}
\label{eqn:style}
\Lstyle = \Vert G(\vggfeat(\tilde{I}_s)) - G(\vggfeat(I_s)) \Vert_2^2,
\end{equation}

\noindent where $G$ computes the Gram matrix of the VGG features extracted from the estimated and ground truth reshaded images. Finally, $\maeWeight$, $\vggWeight$, and $\styleWeight$ define the weight of each term in Eq.~\ref{eqn:total} and we set them to 1e-1, 1e-2, and 1, respectively. Note that we multiply the estimated and ground truth reshaded images by the validity mask before computing each loss term.

\begin{figure}
  \includegraphics[width=\linewidth]{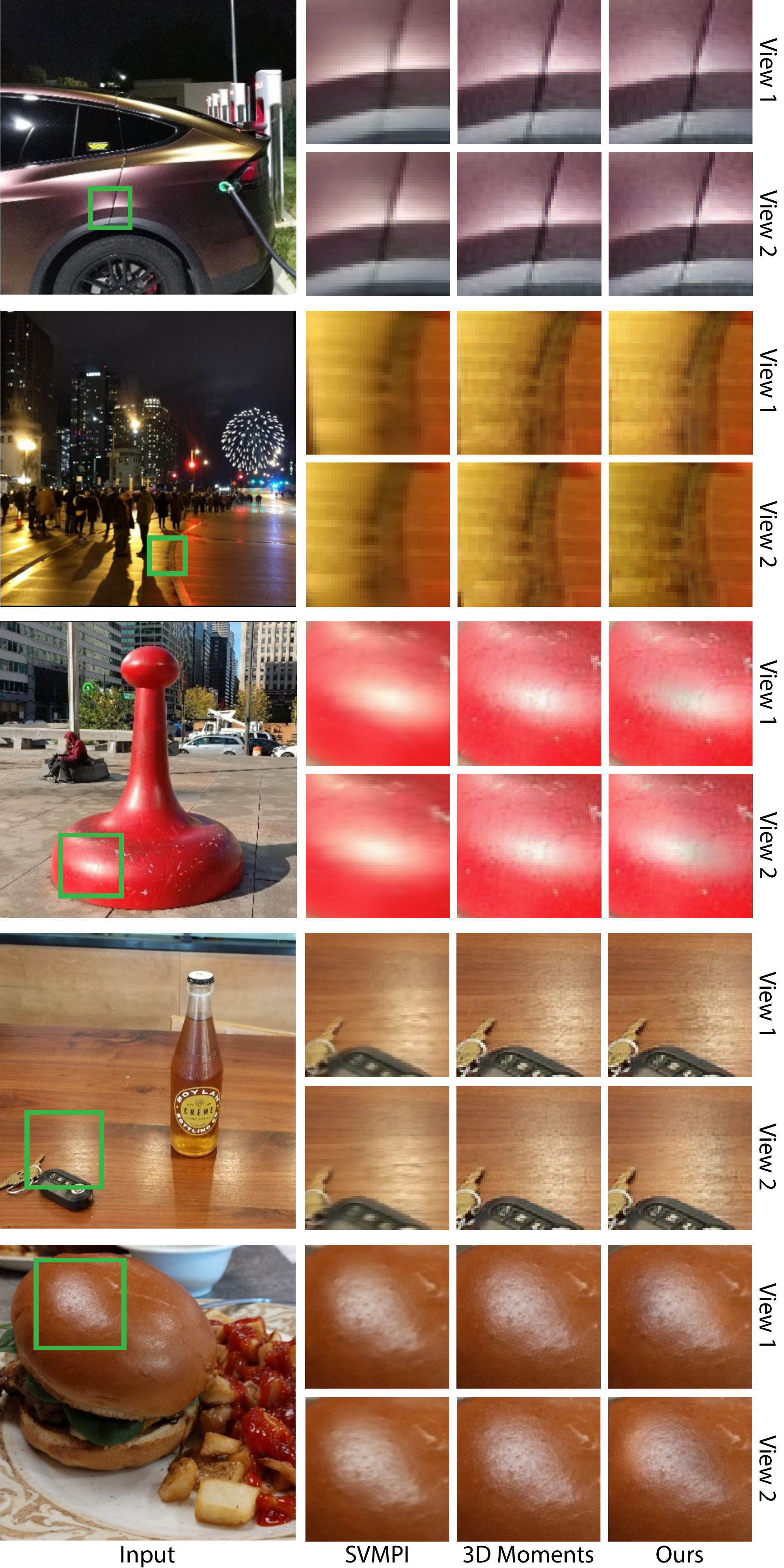}
  \vspace{-0.2in}
  \caption{We show comparisons against SVMPI~\cite{tucker2020mpi} and 3D Moments~\cite{wang2022_3dmoments}. Only our approach is able to move the highlights in different views. Note that we carefully select the insets to cover roughly the same regions in the two views to be able to demonstrate the view dependent effects.}
  \label{fig:results}
  \vspace{-0.25in}
\end{figure}

\subsection{Pixel Relocation}

Once our reshading network is trained, we can use it to reshade the input image during inference. We then use the reshaded image as the input to the approach by Wang et al.~\shortcite{wang2022_3dmoments} to reconstruct the final novel view image. This approach is designed to perform view and time interpolation using near duplicate photos. \revdel{We use our reshaded image as both inputs, since we are only interested in synthesizing views.} \revadd{However, all the operations related to view synthesis utilize a single image. Therefore, we isolate the view synthesis component and use it to generate novel views from a single image.}

The view synthesis component of this approach is an enhanced version of the technique by Shih et al.~\shortcite{Shih3DP20}. Specifically, using the depth, this method first constructs a layered depth image (LDI) representation~\cite{shade1998layereddepth}. It then inpaints the occluded regions and produces LDI features using a network. The LDI features are then warped to the novel view and combined using a subsequent network to produce the final image. Note that our reshaded image is different for each view, which could potentially change the inpainting results, and consequently affect coherency of the synthesized views. However, we did not observe this effect in practice. As shown in the supplementary video, our results are coherent.

\revadd{We note that our approach can be combined with any view synthesis technique that focuses on pixel relocation. We demonstrate this in Table~\ref{tab:one}, where we examine the performance of our approach using Shih et al.'s method~\shortcite{Shih3DP20} (3D Photo) for pixel relocation.}

\section{Results}
\label{sec:results}

We implement our approach in PyTorch and use Adam~\cite{Kingma15Adam} with the default parameters for training. We use a learning rate of 1e-4 for 300K iterations and 1e-5 for another 200K iterations. Our training takes 5 days on an Nvidia 2080 Ti GPU.

We compare our approach against single image view synthesis approaches by Tucker and Snavely~\shortcite{tucker2020mpi} (SVMPI) and Wang et al.~\shortcite{wang2022_3dmoments} (3D Moments). SVMPI is trained in an end-to-end manner on a multi-view dataset and ideally should be able to handle the view-dependent effects. On the other hand, 3D Moments, which we use for pixel relocation, is a modular approach that is not able to move the highlights. We use the code provided by the authors to generate the results. We use images from several datasets, including Holopix50K~\cite{hua2020holopix50k}, Open Images V7~\cite{OpenImages} and Shiny~\cite{Wizadwongsa2021NeX}. Here, we show the image results, but the differences can be better observed in the supplementary video.

\begin{table}%
\caption{\revadd{We show numerical comparisons against the other approaches on three synthetic scenes by evaluating the error between the ground truth and novel view images in terms of PSNR, SSIM, and LPIPS. \revdel{The top and bottom rows show the results for the Teapots and Bathroom scenes, respectively.}}}
\vspace{-0.15in}
\label{tab:one}
\begin{minipage}{\columnwidth}
\begin{center}
\begin{tabular}{llccc}
  \toprule
  Scene & Method & PSNR{$\uparrow$} & SSIM{$\uparrow$} & LPIPS{$\downarrow$}\\
  \toprule
  \midrule
  \multirow{5}{*}{Veach Ajar} & SVMPI & 22.72 & 0.877 & 0.0428\\
  \cmidrule{2-5}
  & 3D Photo & 30.06 & \textbf{0.962} & 0.0200\\
  & 3D Photo + Ours & \textbf{30.70} & \textbf{0.962} & 0.0198\\
  \cmidrule{2-5}
  & 3D Moments & 29.78 & \textbf{0.962} & 0.0149\\
  & 3D Moments + Ours & 30.41 & \textbf{0.962} & \textbf{0.0147}\\
  \midrule
  \multirow{5}{*}{Bathroom} & SVMPI & 20.27 & 0.602 & 0.1255\\
  \cmidrule{2-5}
  & 3D Photo & 29.96 & 0.907 & 0.0329\\
  & 3D Photo + Ours & 30.84 & 0.910 & 0.0323\\
  \cmidrule{2-5}
  & 3D Moments & 32.03 & 0.951 & 0.0284\\
  & 3D Moments + Ours & \textbf{33.12} & \textbf{0.953} & \textbf{0.0281}\\
  \midrule
  \multirow{5}{*}{Modern Hall} & SVMPI & 22.73 & 0.763 & 0.0759\\
  \cmidrule{2-5}
  & 3D Photo & 32.63 & 0.950 & 0.0230\\
  & 3D Photo + Ours & \textbf{32.99} & 0.951 & 0.0229\\
  \cmidrule{2-5}
  & 3D Moments & 30.98 & 0.951 & 0.0197\\
  & 3D Moments + Ours & {31.21} & \textbf{0.953} & \textbf{0.0196}\\
  
  \bottomrule
\end{tabular}
\end{center}
\bigskip\centering
\end{minipage}
\vspace{-0.25in}
\end{table}%

In Fig.~\ref{fig:results}, we show comparisons against the other techniques on five scenes. For each scene, we show the results for two different views. We have carefully selected the insets, so they roughly cover the same region in the two views. Therefore, each approach's ability to adjust the shading based on the view can be observed by comparing the two views. Overall, 3D Moments produce results where the shading of the two views are almost identical. In some cases, SVMPI slightly alters the position of the highlights, but when doing so, it disturbs the texture underneath. Additionally, it produces slightly overblurred results. Our approach, on the other hand, produces detailed images with moving highlights. For example, in the first and fourth rows, our approach moves the highlight to the right and left, respectively, when transitioning from view 1 to 2. Note that our method does not leak the highlights to the dark region in the top row and the diffuse key fob in the fourth row. In the second row, our method produces results with slightly darker shading in the second view, while keeping the underlying texture intact. Finally, in the third and last rows, our approach is able to properly move the highlights (to the left from view 1 to 2) on the red structure and the burger bun, respectively. 

\begin{figure}
  \includegraphics[width=\linewidth]{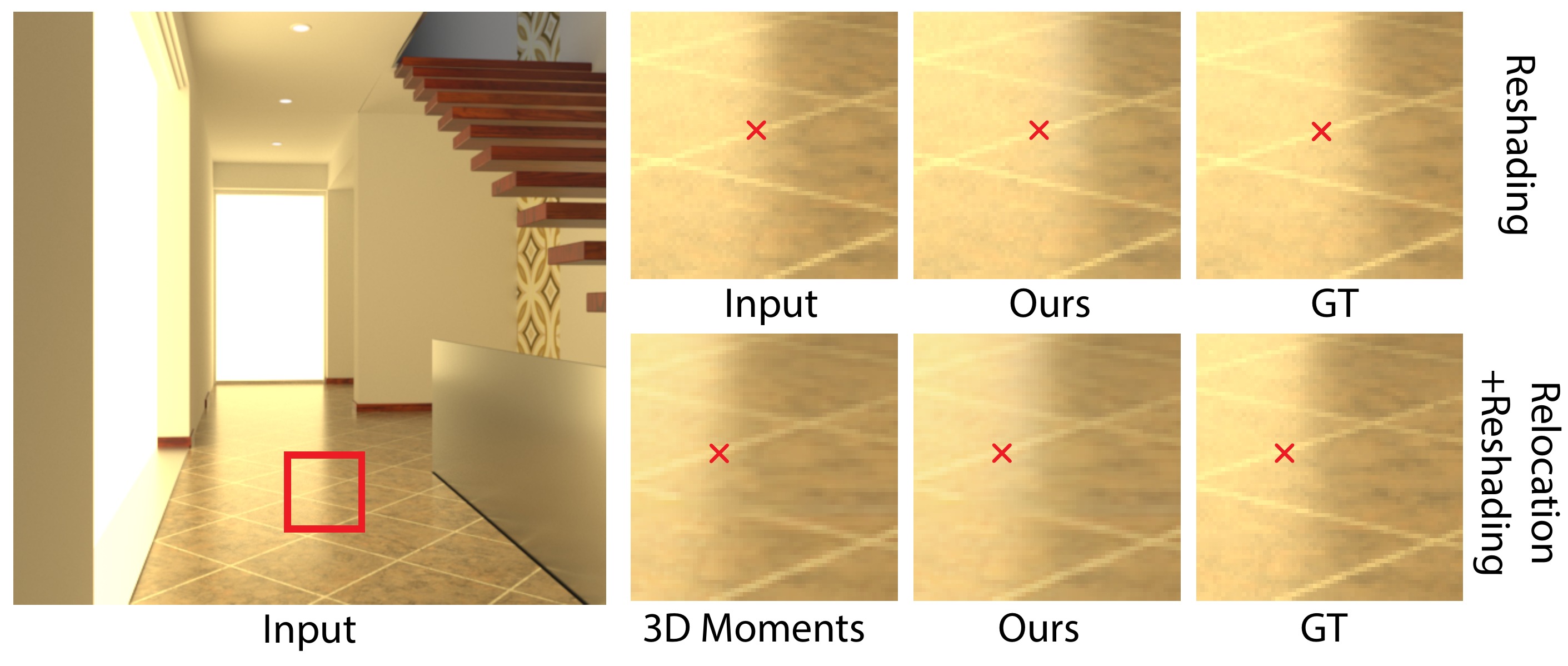}
  \vspace{-0.25in}
  \caption{\revadd{We show our reshading (top) and view synthesis (bottom) results on the Modern Hall scene. Our approach is able to properly move the highlights during the reshading process (top) and produce novel view images that better match the ground truth than existing techniques (bottom).}}
  \label{fig:syn_result}
  \vspace{-0.2in}
\end{figure}

Furthermore, we numerically compare our view synthesis results against the other techniques on three synthetic scenes\revdel{(Teapot and Bathroom on top and bottom, respectively)} in Table~\ref{tab:one}. \revadd{To demonstrate that our approach can be used with any pixel relocation method, we show results with both 3D moments~\cite{wang2022_3dmoments} and Shih et al.'s approach~\shortcite{Shih3DP20} (3D Photo).} \revdel{Overall, our approach produces the best results across all the metrics.}\revadd{As seen, our approach improves the performance of both modular relocation methods.} Note that SSIM and LPIPS are highly sensitive to the textures, but are not sensitive to the smooth highlights. As such, these metrics do not fully reflect our quality improvement. \revadd{Moreover, we evaluate our reshading network in isolation (see Table~\ref{tab:two}), by measuring the error between our synthesized and ground truth reshaded images. By appropriately moving the highlights, our approach produces results that are significantly closer to the ground truth than the input images (without reshading). This is shown visually in Fig.~\ref{fig:syn_result} for the Modern Hall example. Our approach properly moves the highlights (top row), and thus is able to synthesize a novel view image that better matches the ground truth than 3D Moments (bottom row).}

\begin{table}%
\caption{\revadd{We numerically evaluate the effect of reshading in isolation. Our reshading network produces results that are closer to the ground truth than the input.}}
\vspace{-0.15in}
\label{tab:two}
\begin{minipage}{\columnwidth}
\begin{center}
\begin{tabular}{llccc}
  \toprule
  Scene & Method & PSNR{$\uparrow$} & SSIM{$\uparrow$} & LPIPS{$\downarrow$}\\
  \toprule
  \midrule
  \multirow{2}{*}{Veach Ajar} & Input & 35.45 & 0.993 & 0.0012  \\
  & Ours & \textbf{40.10} & \textbf{0.994} & \textbf{0.0008}\\
  \midrule
  \multirow{2}{*}{Bathroom} & Input & 36.50 & 0.991 & 0.0024\\
  & Ours & \textbf{41.20} & \textbf{0.992} & \textbf{0.0020}\\
  \midrule
  \multirow{2}{*}{Modern Hall} & Input & 39.99 & 0.989 & 0.0015\\
  & Ours & \textbf{42.71} & \textbf{0.989} & \textbf{0.0012}\\
  
  \bottomrule
\end{tabular}
\end{center}
\bigskip\centering
\end{minipage}
\vspace{-0.2in}
\end{table}%

\begin{figure}
  \includegraphics[width=\linewidth]{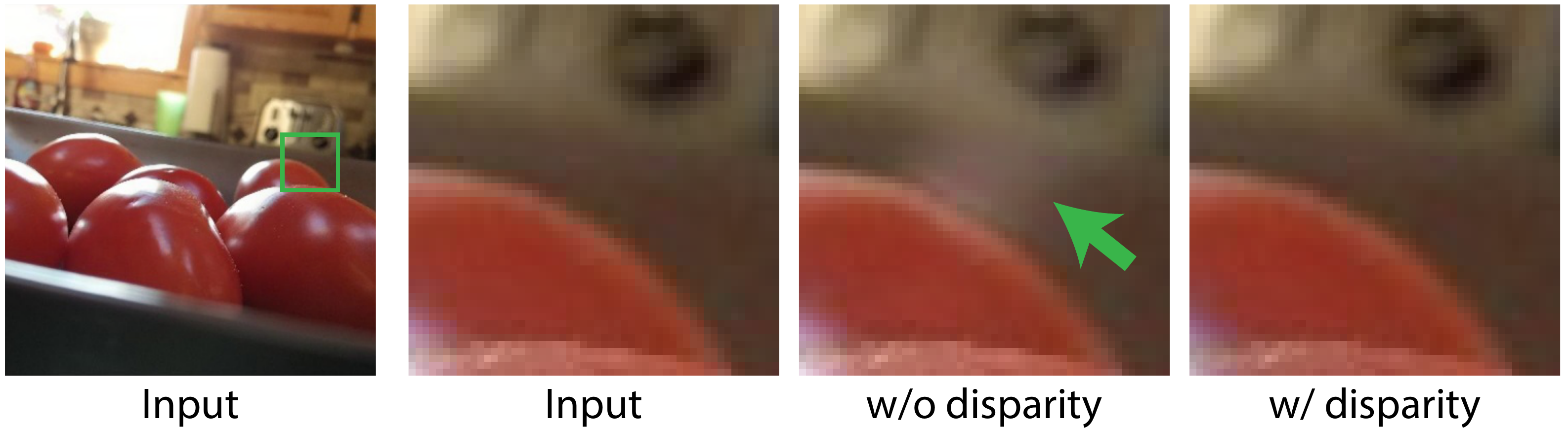}
  \vspace{-0.3in}
  \caption{We evaluate the effect of using disparity as the input to our shading network.}
  \label{fig:ablation_depth}
  \vspace{-0.15in}
\end{figure}

\begin{figure}
  \includegraphics[width=\linewidth]{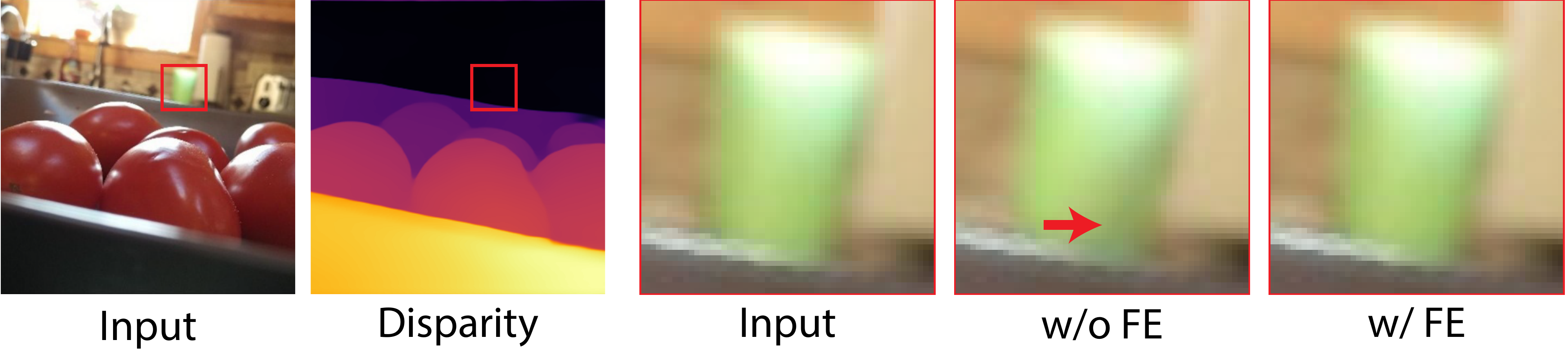}
  \vspace{-0.3in}
  \caption{We compare our results against a version of our approach where we do not apply frequency encoding to the input disparity.}
  \label{fig:ablation_freq_enc}
  \vspace{-0.15in}
\end{figure}

\begin{figure}
  \includegraphics[width=\linewidth]{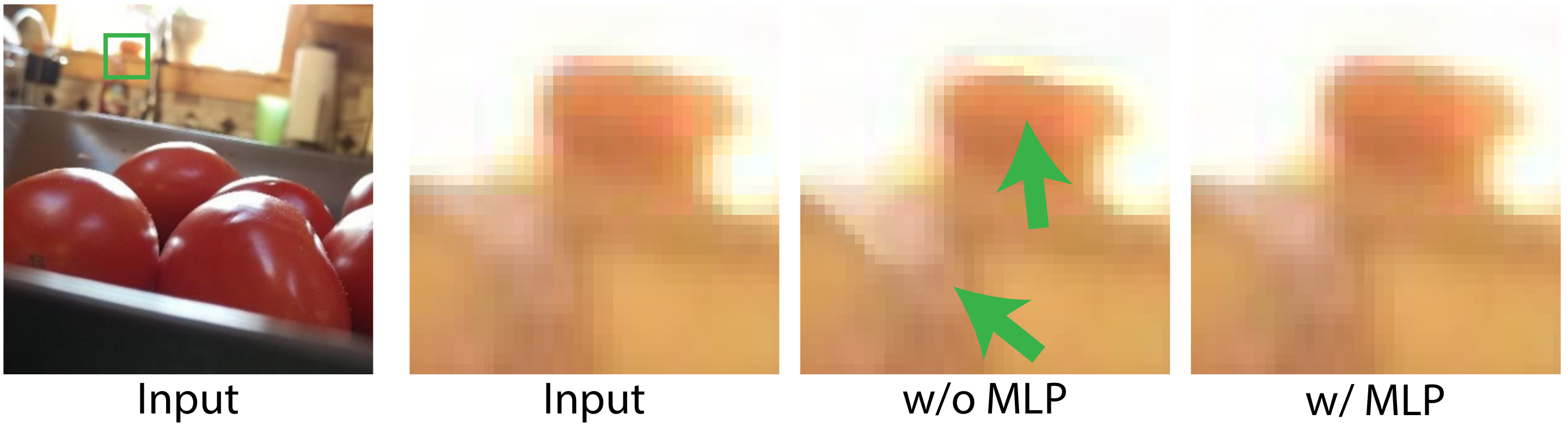}
  \vspace{-0.3in}
  \caption{We compare our results against a version of our approach where the camera position is directly concatenated to the bottleneck features of the UNet.}
  \label{fig:ablation_mlp}
  \vspace{-0.15in}
\end{figure}

Next, we discuss the effect of several design choices in our approach numerically (Table~\ref{tab:three}) and visually (Figs.~~\ref{fig:ablation_depth},~\ref{fig:ablation_freq_enc},~and~\ref{fig:ablation_mlp}). In Fig.~\ref{fig:ablation_depth}, we demonstrate that without the disparity as the input, our reshading network is not able to detect the depth discontinuities and smears the shading of the tomato on the bowl. Moreover, as shown in Fig.~\ref{fig:ablation_freq_enc}, without frequency encoding, our network has difficulty handling the objects that are far away and incorrectly changes their shading. Finally, in Fig.~\ref{fig:ablation_mlp} we show the result of directly concatenating the camera pose with the bottleneck features (w/o MLP). As seen, without the MLP, our network cannot effectively utilize the camera information and incorrectly changes the shading of the background areas.

\begin{table}%
\caption{We show numerical comparisons against variations of our approach without disparity, frequency encoding, and MLP. \revadd{The results are averaged over the three synthetic scenes}. As shown, all these components are necessary to achieve the best results.}
\vspace{-0.15in}
\label{tab:three}
\begin{minipage}{\columnwidth}
\begin{center}
\begin{tabular}{lccc}
  \toprule
  Method & PSNR\revadd{$\uparrow$} & SSIM\revadd{$\uparrow$} & LPIPS\revadd{$\downarrow$}\\
  \toprule
  \midrule
  w/o disparity & 30.98 & 0.950 & 0.0213\\
  w/o FE & 31.11 & 0.950 & 0.0210 \\
  w/o MLP & 31.17 & 0.951 & 0.0211\\
  Ours & \textbf{31.58} & \textbf{0.956} & \textbf{0.0208}\\
  
  \bottomrule
\end{tabular}
\end{center}
\bigskip\centering
\end{minipage}
\vspace{-0.25in}
\end{table}%

\section{Limitations}
\label{sec:discussion}

\begin{figure}
  \includegraphics[width=\linewidth]{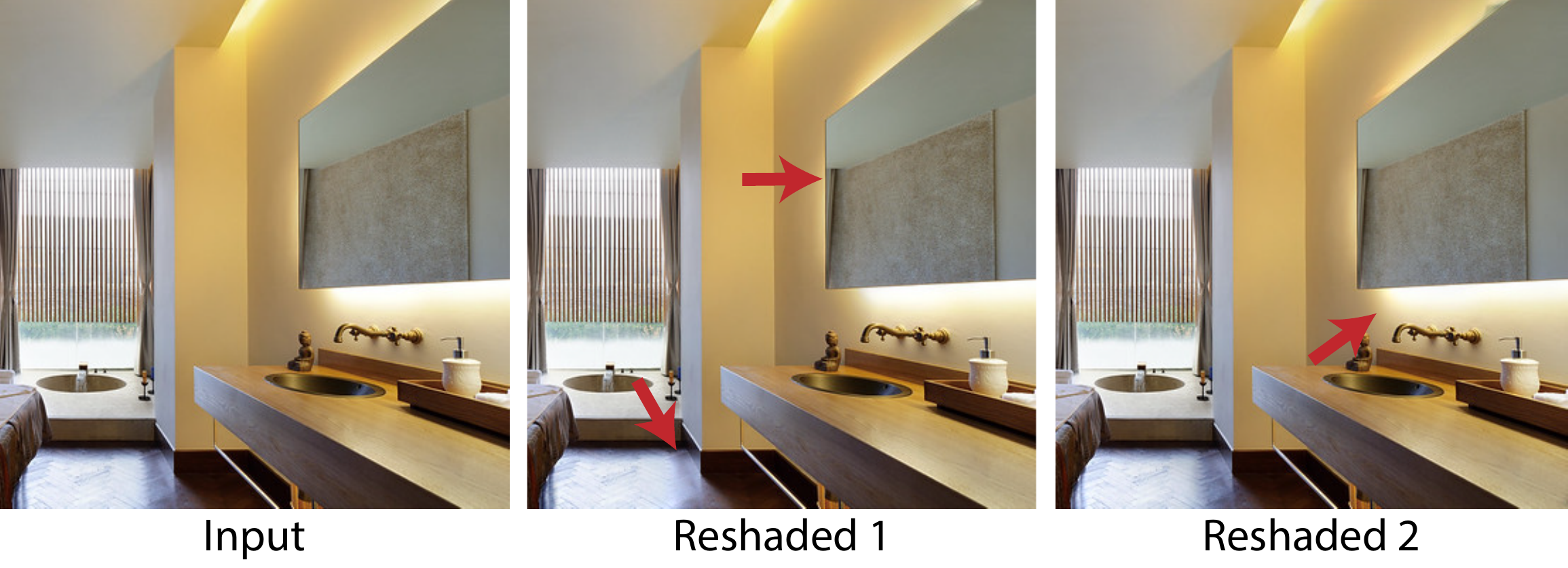}
  \vspace{-0.2in}
  \caption{We show the input image as well as two reshaded images corresponding to different views. As seen, our method is not able to properly move the content of the mirror on the right wall in the two reshaded images. Additionally, while our method correctly moves the highlights on the ground, it detects the strong saturated regions under the mirror as highlights and move them in the reshaded images.}
  \label{fig:lim_mirror}
  \vspace{-0.2in}
\end{figure}

Although we have demonstrated that our simple network can produce reasonable results, this is an extremely challenging problem and, as shown in Fig.~\ref{fig:lim_mirror}, our approach has several limitations. For example, we are currently not able to handle highly specular surfaces, such as mirrors. As shown in Fig.~\ref{fig:lim_mirror} (mirror on the right wall), our technique is not able to correctly move the content inside the mirror between the two reshaded images. Additionally, in cases where the light sources are very close to diffuse surfaces, they create strong saturated regions (see the area underneath the mirror). In these cases, our reshading network interpret these as highlights and moves them between different views. 

\section{Conclusion}
We have presented a method to handle view dependent effects in single image novel view synthesis. Specifically, we propose to split the task of view synthesis into pixel reshading and relocation processes and treat them independently. We use a network to adjust 
the shading of the input image according to the novel camera. We then use the reshaded image as the input to an existing view synthesis method to perform the pixel relocation task. We demonstrate that our method produces plausible results with view-dependent highlights that are better than the existing methods.

\begin{acks}
The authors would like to thank the reviewers for their comments and suggestions. This work was funded by Leia Inc. (contract \#415290). Nima Khademi Kalantari was in part supported by CAREER Award (\#2238193). Portions of this research were conducted with the advanced computing resources provided by Texas A\&M High Performance Research Computing.
\end{acks}

\bibliographystyle{ACM-Reference-Format}
\bibliography{sample-bibliography,References}

\end{document}